\documentclass{article}
\usepackage{ijcai20}

\usepackage{soul}
\usepackage{url}
\usepackage[hidelinks]{hyperref}
\usepackage[utf8]{inputenc}
\usepackage[small]{caption}
\usepackage{graphicx}
\usepackage{amsmath}
\usepackage{booktabs}
\usepackage{amsthm,amssymb,amsfonts}
\usepackage{tikz,lipsum,lmodern}
\usepackage[most]{tcolorbox}
\usepackage{times}

\newcommand{\timex}{timex}

\title{A Survey on Temporal Reasoning for \\Temporal Information Extraction from Text \\(Extended Abstract)\footnote{This paper is an extended abstract of an article in the Journal of Artificial Intelligence Research [Leeuwenberg and Moens, 2019].}}

\author{
Artuur Leeuwenberg$^1$\And
Marie-Francine Moens$^2$\\
\affiliations
$^1$Julius Center, University Medical Center Utrecht, The Netherlands\\
$^2$Department of Computer Science, KU Leuven, Belgium\\
 \vspace{.1cm}aleeuw15@umcutrecht.nl, sien.moens@cs.kuleuven.be
}

\begin{document}

\maketitle

\begin{abstract}
Time is deeply woven into how people perceive, and communicate about the world. Almost unconsciously, we provide our language utterances with temporal cues, like verb tenses, and we can hardly produce sentences without such cues. 
Extracting temporal cues from text, and constructing a global temporal view about the order of described events is a major challenge of automatic natural language understanding. Temporal reasoning, the process of combining different temporal cues into a coherent temporal view, plays a central role in temporal information extraction. This article presents a comprehensive survey of the research from the past decades on temporal reasoning for automatic temporal information extraction from text, providing a case study on the integration of symbolic reasoning with machine learning-based information extraction systems.
\end{abstract}

\vspace{-.5cm}
\section{Introduction}
Human language is filled with cues about the timing of the events that we communicate about.
\textbf{T}emporal \textbf{I}nformation \textbf{E}xtraction (TIE) is the process of automatically extracting temporal cues from text, with the goal to construct a (possibly underspecified) timeline of events from them.

TIE not only plays a major role in the general problem of natural language understanding (NLU), but is also used in many applications: information retrieval \cite{campos2015survey}, question answering \cite{meng2017temporal}, and multi-document summarization \cite{ng2014exploiting}, and has great potential in the clinical domain, for applications like patient timeline visualization \cite{jung2011building}, forecasting treatment effects \cite{zhou2007temporal}, and patient selection for clinical trials \cite{raghavan2014essential}. 
Because of the strong structure of time, and the great variation in the types of temporal cues we can express in language, a central challenge in TIE is how to combine all these separate cues into a single coherent temporal ordering of the described events. To achieve this, temporal reasoning is of vital importance.
We define \textbf{T}emporal \textbf{R}easoning (TR) in the context of TIE as the process of combining different (annotated or extracted) temporal cues to derive additional temporal information about the text. TR is crucial for TIE and has already been exploited widely in the research community in all stages of TIE model construction: annotation, pre-processing, model training, inference, and evaluation.

To cover the evolution of TR approaches for TIE in parallel with the popularization of using machine learning (ML) methods for natural language processing (NLP), the focus of this survey lies on the research on TR for TIE systems from the past three decades. We abstract from what linguistic features are successful for TIE systems, as this is discussed in depth by \cite{derczynski2017automatically}, and can be considered complementary to the focus of the survey. This extended abstract aims to capture the most important elements and conclusions from the original article \cite{leeuwenberg2019survey}.

In the context of the interesting and important new developments in the past years this article provides the following contributions:
\begin{itemize}
\item An introduction to the various types of temporal information in language, and a brief explanation introducing the background on TR for TIE required for the latest state-of-the-art TIE models.
\item An overview of the various ways in which TR has been exploited in TIE models over the past decades.
\item A distillation of the most important conclusions to successfully incorporate TR in a TIE system and some promising directions for future work.
\end{itemize}

\nocite{freksa1992temporal}

\begin{figure*}[h!]
    \centering
    \vspace{-.1cm}
    \includegraphics[width=\textwidth]{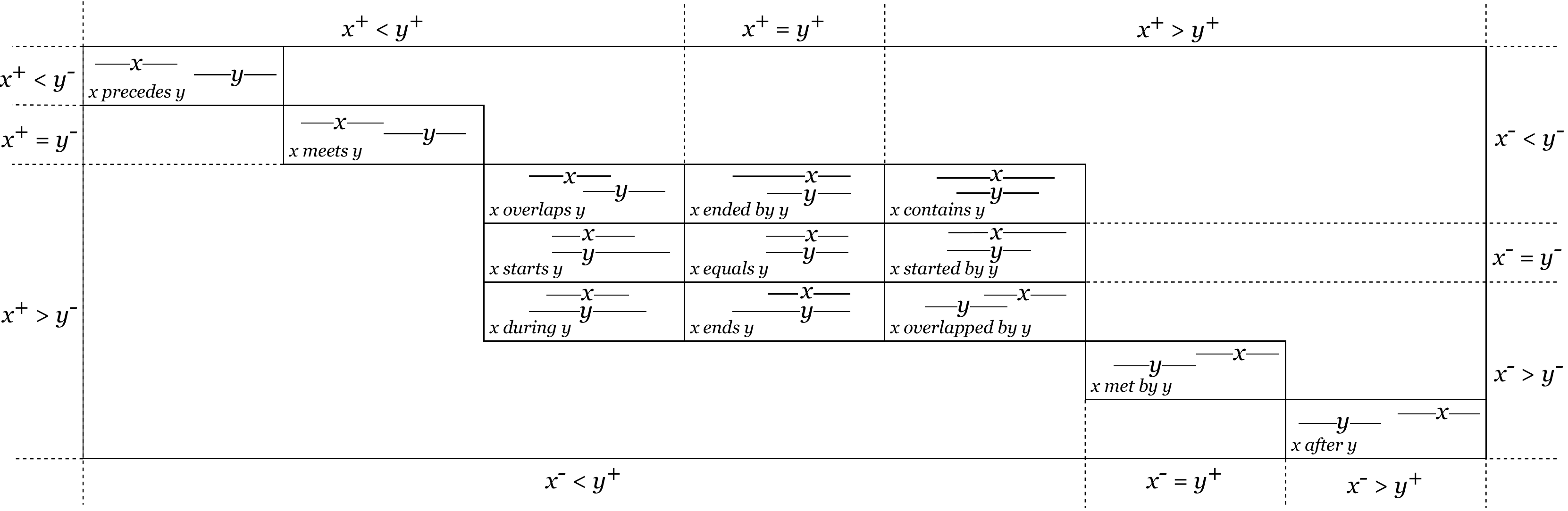}
    \vspace{-.3cm}
    \caption{A lattice [Freksa, 1992] showing Allen's 13 basic \textit{interval} relations (small boxes), and their mapping to \textit{point}-algebraic constraints (outside the large rectangle). Start and end points of intervals $x$ and $y$ are noted by $x^-, x^ +$, and $y^-, y^+$ respectively. Conceptual neighborhood between interval relations is proportional to Manhattan distance [Leeuwenberg and Moens, 2019].}
    \label{fig:lattice}
    \vspace{-.4cm}
\end{figure*}
\vspace{-.3cm}
\section{Temporal Cues in Language}

It is important to study the types of temporal information that can be expressed in language because the different cues need to be combined by a TR system in order to build a complete temporal overview of the text. The most common interpretation of events on the timeline is as an interval (some expressions may refer to sets of events/intervals, or very short almost instantly occurring events). The span of the interval corresponds to the time that the event takes place. In this section, we give a short exemplified overview of different temporal cues in language to show what types of temporal information the cues can provide, i.e., in what way the cues may constrain the position of event intervals on the timeline. 

Temporal cues can refer to various components of the event intervals: the entire \textit{interval}, but also just its \textit{start}, \textit{end}, or \textit{duration}. For instance, in Ex. 1, the \textit{duration} of the antibiotics administration is given (\textit{10 days}), and so is its \textit{start time} (sometime on the 2$_{
\text{nd}}$ of June). However, for the improvement in respiratory status only the \textit{end time} is mentioned explicitly (last 2 days of the antibiotics administration).\vspace{.2cm}\\
\textbf{Ex. 1} \textit{Antibiotics were started on 6/2 and continued for 10 days. Respiratory status improved up til the last 2 days.}
\vspace{.2cm}

The type of information given for each component can also differ: it can be \textit{absolute} (e.g., in Ex. 1, the duration of the antibiotics of \textit{10 days}), or \textit{relative} to other intervals (e.g., in Ex. 2 the start time of the third set is \textit{after} the first two). Additionally, the information can be \textit{definite}, with clear interpretable timings (e.g., in Ex. 2, the duration of \textit{28 minutes}), or indefinite, using \textit{vague} quantification (in Ex. 2, the duration of the first two sets, each \textit{nearly an hour}).\vspace{.2cm}\\
\textbf{Ex. 2} \textit{After the grueling duels of the first two sets, which each had taken nearly an hour, Nadal won the third set in 28 minutes.}
\vspace{.1cm}

Even with all these explicit cues, \textit{world knowledge} about the typical order of events, their durations, and even typical time between events may be needed to properly interpret the temporal meaning that is to be conveyed (e.g., although not explicitly mentioned, one would assume that the third set took place on the same day as the first two). Finally, temporal information provided through language is generally \textit{underspecified} (e.g., based on the example texts, we don't know in what year any of the events took place). A good TR system should be able to flexibly incorporate all these different types of information, and possibly their uncertainty, to come to the timeline that is to be conveyed by the writer.
  \vspace{-.1cm}

\section{Reasoning with Intervals and Points}
\begin{figure}[]
 
\centering
\includegraphics[]{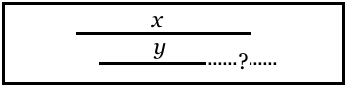}
\caption{\label{fig:indefinite} General Allen relation: $x$ \{contains, overlaps, ended by\} $y$, described by the text "$y$ started during $x$".}
    \vspace{-.3cm}
\end{figure}

The most popular TR framework used in TIE is Allen's interval algebra (AIA) \cite{allen1990maintaining}. AIA is based on a set of thirteen mutually exclusive basic interval relations (shown in Fig. \ref{fig:lattice}) that can be assigned to any pair of \textit{definite} intervals (where relative positions of start and end-points are known). If for two intervals $x$ and $y$ information about the relative positions of their start and end-points is incomplete (i.e., their relation is indefinite, as in Fig. \ref{fig:indefinite}) a \textit{general} Allen relation can be assigned, expressed as the disjunction of all possible basic relations, given the information provided. The full AIA covers the powerset of all thirteen basic relations, comprising $2^{13}=8192$ relations.

To infer new relations from a graph of general Allen relations between different events, a composition table is used, containing transitivity rules for all basic relations, i.e., it shows for any pair of connected relations $r_1(x, y)$ and $r_2(y, z)$, what relation $r_3(x, z)$ could be inferred. For the full AIA, computing all inferrable relations (taking the transitive closure) is, like checking consistency, NP-complete \cite{vilain1990constraint}, making full AIA impractical for TIE systems. Many TIE systems that use TR are limited to using basic relations (i.e., definite relations). Although hardly used in TIE systems, many tractable sub-fragments of AIA exist \cite{krokhin2003reasoning}.
Some of which focus on integrating quantitative reasoning \cite{meiri1996combining} or uncertainty \cite{schockaert2008temporal}. A major insight on which many efficient TR algorithms are built is the fact that the basic Allen relations can be expressed as sets of point-algebraic relations ($<, =, >$), as shown in Fig. \ref{fig:lattice}. Point algebra (PA) can express each basic Allen relation as a \textit{conjunctive} set of point-wise relations between the starts and endings of the intervals. For example, $x$ \{equals\} $y$ can be expressed by the conjunctive set $\{x^-=y^-, x^+=y^+\}$, i.e., if intervals $x$ and $y$ are equal, then their start and endings are also equal (and vice-versa). They also allow expression of many indefinite relations (e.g., the one from Fig. \ref{fig:indefinite}, as $\{x^- < y^-,   x^+ > y^-\}$). Reasoning in PA requires a much smaller transitivity table, which contributes to its efficiency, important for practical TR systems.

  \vspace{-.2cm}
\section{Annotation Schemes}
The scheme for temporal annotation in text steers the types of temporal cues that can be extracted by TIE systems, and consequently how TR can be exploited.

The most popular scheme is TimeML \cite{pustejovsky2003timeml}, used as the basis for many corpora, and for TempEval, a series of shared tasks on TIE evaluation that resulted in many of the existing TIE systems \cite{verhagen2007semeval}.
The core TimeML concepts are event expressions and time expressions (timex). Event expressions refer to events in the real world, and \timex{} refer to calendar dates, times, and definite durations, among others. In TimeML, events and \timex{} expressions are temporally connected through interval relations, called temporal links (TLinks), approximately following Allen's basic interval relations, consequently focusing on definite information. To go beyond the focus on explicit and definite information in TimeML, many interesting proposals have been made: to include a \textit{vague} relation to annotate indefinite relations \cite{cassidy2014annotation}, to annotate point-wise relations between start and end points of events instead of basic interval relations \cite{ning2018multi}, annotation of durations for all events (not only for events with an explicit definite duration \timex{} as in TimeML) \cite{pan2006annotated}, and to annotate temporal position of events on the absolute calendar instead of by interval relations \cite{reimers2016temporal}, as pair-wise temporal relation annotation is a difficult task for annotators \cite{derczynski2016representation}. These proposals provide new opportunities for further exploitation of TR.

\section{Survey of Current Approaches}
This section desribes how TR has been incorporated in existing TIE systems throughout the steps of model development: annotation, data pre-processing, training and prediction, and evaluation. This abstract focuses purely on the task of temporal relation extraction among events and \timex{}, as TR plays a large role in this task, and performance metrics are generally lower compared to event and \timex{} extraction.

\subsection{Annotation \& Data Preprocessing}
A widely recognized problem in temporal annotation, is that annotators regularly miss temporal cues in the text, especially when annotating TLinks between pairs of events, as there are so many possible relations to annotate. \cite{verhagen2005temporal} argue that not all event-pairs are equally useful to annotate, as some can be inferred with TR from earlier annotations. Another motivation to exploit TR during annotation is that annotators sometimes annotate temporally inconsistent graphs, which impedes training of TIE systems.

A frequently made TR-based preprocessing step is to calculate the transitive closure over the annotated relations, to obtain a more dense graph. The effect of temporal closure on performance for models that use no TR during prediction is ambivalent. It may improve performance \cite{mani2006machine}, or decrease performance \cite{tatu2008experiments}. However, when using a model that exploits TR also during prediction, data expansion through a temporal closure can be beneficial \cite{chambers2008jointly}. 

\subsection{Training and Prediction}
To enhance temporal relation extraction models, TR can be integrated during training, and prediction. There are various ways how integration of TR in training and prediction can improve TIE models: (1) ensure temporal consistency among predicted relations, and (2) to constrain the output space and improve prediction accuracy.

\paragraph{Greedy Inference.} Two greedy approaches have been proposed to construct a graph of temporal relations: relations are added one-by-one, in a certain order, checking temporal consistency after each addition. The order can be natural reading order, or follow the score from a local classifier (best first). Both approaches improved results over absence of TR, the last surpassing the first \cite{bramsen2006inducing}.

\paragraph{Post-Hoc Conflict Resolution.}
Another approach is to first predict the full graph with a local model, and detect and resolve inconsistencies afterwards, by removing conflicting edges based on confidence \cite{verhagen2008temporal,tatu2008experiments} among others. Most works report a positive impact of this method, but differ in how they remove edges. It could be interesting to compare these to each other in more detail.

\paragraph{Sieve-Level and Stacked Inference.}
The sieve-based method \cite{chambers2014dense} is flexible in incorporation of both rule-based and machine learning components. This approach extract TLinks in consecutive phases by separate model components (or sieves). Each sieve extracts TLinks, using the original input text, and the outputs from earlier sieves. TR is incorporated by taking a transitive closure after applying each sieve. This prevents later sieves from assigning TLinks that are inconsistent with those extracted by earlier sieves.
Similar approaches have been adopted by \cite{mcdowell2017event}, and \cite{mirza2016catena}, who report an increase in recall (and F$_1$) due to sieve-level TR.

Much alike are methods that stack learners on top of each other to correct mistakes from earlier stages \cite{laokulrat2015stacking}, which can even be trained jointly across stacks \cite{meng2018context}.

\paragraph{Markov Logic Networks.}
Another important method for TR-based prediction are Markov Logic Networks (MLN), first explored for TIE by \cite{yoshikawa2009jointly}. A major difference between MLN and the methods mentioned earlier, is that instead of combining locally trained models in a global inference setting, MLN also incorporate the temporal constraints in training. The weights for TR constraints allow the model to learn soft correlations between TLinks. Model different transitivity rules with MLN, \cite{yoshikawa2009jointly} show that incorporating TR during training and prediction outperforms local models.

\paragraph{Integer Linear Programming.}
Where MLN use soft constraints, integer linear programming (ILP), a method for constrained optimization, uses hard constraints to cut off large areas from the search space, which can result in faster inference \cite{mojica2016markov}. ILP for TR-based prediction was first proposed by \cite{bramsen2006inducing}, to maximize the sum of pair-wise classifier scores under TR-based constraints. ILP has been adopted widely and successfully for TIE. In contrast to many, \cite{denis2011predicting} formulated their ILP objective in PA for more efficient TR. Recent approaches not only use ILP during prediction, but also during training, combining ILP with a structured perceptron \cite{leeuwenberg2017structured,ning2017structured}.

\paragraph{Direct Timeline Models.}
Recently, a new approach to temporal event ordering was proposed \cite{leeuwenberg2018timeline}. Instead of predicting TLinks among events and temporal expressions, their model predicts start and end points of events such that their order satisfies the annotated TLinks. Prediction time of this approach is linear in the number of events, in contrast to predicting a graph of TLinks, which is quadratic or requires pruning. To train their model they exploit TR to convert TLinks to sets of PA constraints.

\subsection{Evaluation}
A problem in evaluation of TIE, is that the same timeline can be represented by different sets of relations: if a system predicts \textsc{before}($x$, $y$) it should not be penalized if the ground truth is \textsc{after}($y$, $x$). For this, taking a temporal closure over the predicted and ground truth graphs has been proposed, in combination with different ways to weigh original vs. inferred relations \cite{setzer2003using,tannier2011evaluating,UzZaman:2011}, of which the \textit{temporal awareness} metric \cite{UzZaman:2011} is the most widely used evaluation metric.

\section{Conclusions and Future Work}

In this abstract, we provide a short demonstration of various types of temporal information in language, briefly introduced the most frequently used framework in TR for TIE, and reviewed different methods to exploit TR for constructing TIE models, distilling the most widely confirmed conclusions. 

In closing, it is clear that TR is crucial for TIE, and widely used in all aspects of model construction: annotation, data preprocessing, training, prediction, and evaluation. However, most current research on TIE still addresses sub-fragments of the complete TIE problem, focusing on extraction of specific types of temporal cues, instead of extracting all cues jointly which would allow them to complement each other. Consequently, it remains an open research question how to perform efficient and expressive TR involving all types of temporal cues. We believe to answer this question, a flexible, expressive and efficient reasoning framework is required. For this, we believe important directions of research are point-based reasoning approaches, striking a good balance between efficiency and expressiveness, and (deep) machine learning methods, to facilitate flexibility in model construction, multi-task learning, and sharing of representations.

\section*{Acknowledgements}
This work was supported
by the ERC Advanced Grant CALCULUS (788506), and by
the IWT-SBO project ACCUMULATE (150056).

\bibliographystyle{ijcai20}
\bibliography{ijcai20}
\end{document}